\documentclass[conference]{IEEEtran}
\IEEEoverridecommandlockouts
\usepackage{amsmath,amssymb,amsfonts}
\usepackage{graphicx}
\usepackage{epstopdf}
\usepackage{textcomp}
\usepackage{bm}
\usepackage{algorithm}
\usepackage{algorithmic}
\usepackage{multirow}
\usepackage{array}
\usepackage{hyperref}
\usepackage{makecell}
\usepackage{balance}
\usepackage{units}
\usepackage{siunitx}
\usepackage{xcolor}

\def\BibTeX{{\rm B\kern-.05em{\sc i\kern-.025em b}\kern-.08em
    T\kern-.1667em\lower.7ex\hbox{E}\kern-.125emX}}
\begin{document}

\title{GGT: Graph-Guided Testing for Adversarial Sample Detection of Deep Neural Network
}

\author{
    \IEEEauthorblockN{Zuohui Chen$^{1*}$, Renxuan Wang$^{1*}$, Jingyang Xiang$^1$,  Yue Yu$^2$, Xin Xia$^3$, Shouling Ji$^4$, \\
    Qi Xuan$^{1{\dag}}$, Xiaoniu Yang$^1$}
    \IEEEauthorblockA{$^1$ Institute of Cyberspace Security, Zhejiang University of Technology, Hangzhou, 310023, China}
    \IEEEauthorblockA{$^2$ National University of Defense Technology, Changsha, 410073, China}
    \IEEEauthorblockA{$^3$ Monash University, Melbourne, Australia}
    \IEEEauthorblockA{$^4$ Zhejiang University, Hangzhou, 310023, China}
    \IEEEauthorblockA{\{zuohuic\}@qq.com, \{2111903087\}@zjut.edu.cn, \{xiangxiangjingyang\}@gmail.com, \{yuyue\}@nudt.edu.cn, \\
    \{xin.xia\}@monash.edu, \{sji\}@zju.edu.cn, \{xuanqi\}@zjut.edu.cn}
}


\maketitle
\begin{abstract}
Deep Neural Networks (DNN) are known to be vulnerable to adversarial samples, the detection of which is crucial for the wide application of these DNN models. Recently, a number of deep testing methods in software engineering were proposed to find the vulnerability of DNN systems, and one of them, i.e., Model Mutation Testing (MMT), was used to successfully detect various adversarial samples generated by different kinds of adversarial attacks. However, the mutated models in MMT are always huge in number (e.g., over 100 models) and lack diversity (e.g., can be easily circumvented by high-confidence adversarial samples), which makes it less efficient in real applications and less effective in detecting high-confidence adversarial samples. In this study, we propose Graph-Guided Testing (GGT) for adversarial sample detection to overcome these aforementioned challenges. GGT generates pruned models with the guide of graph characteristics, each of them has only about 5\% parameters of the mutated model in MMT, and graph guided models have higher diversity. The experiments on CIFAR10 and SVHN validate that GGT performs much better than MMT with respect to both effectiveness and efficiency.
\end{abstract}




\section{Introduction}
\label{Sec:Intro}
Deep Neural Networks (DNN) have been widely used in many applications, e.g., autopilot~\cite{bojarski2016end}, speech recognition~\cite{carlini2018audio}, and face recognition~\cite{athalye2018synthesizing}. For certain tasks, its capability is even better than humans, making it an indispensable part of some critical systems, such as self-driving cars~\cite{bojarski2016end}, access control systems~\cite{aneja2018iot}, and radio systems~\cite{riyaz2018deep}. However, the safety of DNN has been widely concerned, i.e., it is vulnerable to adversarial samples, which were first discovered by Szegedy et al.~\cite{szegedy2013intriguing} and means a kind of samples formed by adding a small perturbation to natural samples. The tiny changes of samples usually do not affect human judgment, but it can indeed make the trained model return a different output from the original sample with high confidence.

The researchers of the Machine Learning (ML) community have been studying the relationship between adversarial samples and DNN trustworthiness~\cite{sun2019towards}. From the perspective of algorithm, it is widely considered that the existence of adversarial samples is caused by the extremely high dimension of input feature space and the linear nature of the DNN model~\cite{goodfellow2014explaining}; while other researchers may also argue that adversarial samples are not the problem of model structure, but rather the features of datasets~\cite{ilyas2019adversarial}. The Software Engineering (SE) community, on the other hand, considers the DNN model as a kind of software and adversarial samples as the bugs hidden inside the program. To detect potential defects of a DNN model, various testing methods were proposed to perform quantitative analysis on the quality of DNN models. Neural coverage~\cite{tian2018deeptest} is the most used and the first criteria for testing. After that, more and more testing methods are emerging, e.g., surprise adequacy~\cite{kim2019guiding}, mutation testing~\cite{ma2018deepmutation}, neuron boundary coverage~\cite{ma2018deepgauge}, DeepXplore~\cite{pei2017deepxplore}, DeepInspect~\cite{tian2020testing}, fuzzing testing~\cite{guo2018dlfuzz}, and the integrated testing framework~\cite{xie2019deephunter}.


Several DNN testing methods were validated to be effective in detecting adversarial samples as a kind of bug in DNN systems. For instance, Wang et al.~\cite{wang2019adversarial} used Model Mutation Testing (MMT) to detect adversarial samples; Kim et al.~\cite{kim2019guiding} showed that adversarial samples and normal samples have different surprise adequacy; Wang et al.~\cite{wang2020dissector} found that adversarial samples can be distinguished with the model hidden layer output. However, adversarial sample detection mechanisms may be still vulnerable to adaptive attacks, i.e., the attacker grabs the model information and defense strategy. Tramer et al. evaluated~\cite{tramer2020adaptive} the most recent works on adversarial defense and detection proposed by the ML community. They found that almost all of these methods can be circumvented with their accuracy substantially reduced from what was originally claimed. One possible reason is that most of the evaluated methods are based on a single model, and thus are very easy to be targeted.


MMT is one of the state-of-the-art \emph{multiple-model} method on detecting adversarial samples~\cite{wang2019adversarial}. The idea is that adversarial samples are more sensitive to the slight change of the decision boundary. It first creates a group of mutated models and then detects adversarial samples based on the consistency of the outputs generated by the mutated models and the original model. Though MMT achieved great success in adversarial sample detection, there are two main disadvantages that may hinder its wide application in reality. First, it may need over 100 mutated DNN models (with a similar computational cost to the original model) to detect adversarial samples, leading to an unacceptable cost in practical use~\cite{wang2020dissector}, especially on the embedded systems with limited computational and storage resources. Second, since the mutated models have very similar decision boundaries to the original model, such lack of diversity could make MMT fail to detect high-confidence adversarial samples relatively far from the decision boundary.

To overcome the above shortcomings, in this paper, we propose Graph-Guided Testing (GGT) for adversarial detection, which is also a multiple-model method. We argue that DNN of a certain structure can be beneficial to image classification, as well as adversarial detection, while the structure of DNN can be designed or optimized by the theory in network science. The relational graph is the most recently proposed technology to facilitate the design of DNN structure. According to the characteristics of the relational graph, we selectively remove some of the edges in the graph to achieve the pruning of DNN and obtain diverse decision boundaries. We also exploit the average number of edges connected to a node, i.e., Average Degree (AD), and the average distance between any two nodes in the network, i.e., Average Shortest Path Length (ASPL), to obtain more competitive detection results. By comparing with MMT, the generated DNN models in our GGT have much less floating point operations (FLOPs), about only 5\% of the original model, and meanwhile, the high diversity of DNN structure leads to significantly better performance on detecting adversarial samples with an even smaller number of models required. In particular, we make the following contributions.
\begin{enumerate}
\item We propose a graph-guided pruning method for DNN models to significantly reduce the model computational cost with low accuracy loss, where the Average Degree (AD) of the graph is used to control the number of FLOPs in the generated DNN model.
\item We propose Graph-Guided Testing (GGT) for adversarial detection, which can detect adversarial samples more efficiently, i.e., using significantly fewer additional models with much less computational cost, by comparing with MMT.
\item We find that the Average Shortest Path Length (ASPL) is correlated with both the accuracy of individual models and the adversarial detection performance of GGT. The graph with shorter ASPL can naturally generate a DNN model with a higher accuracy, which can be further used to better distinguish adversarial samples and normal samples, and thus is more suitable to establish our GGT.
\item We compare our GGT with MMT on CIFAR10 and SVHN, and find that GGT indeed outperforms MMT both on accuracy and efficiency. MMT achieves the average accuracy of 74.74\% and 44.79\% on CIFAR10 and SVHN by using 51.12 and 77.16 models on average, respectively; while GGT detects  93.61\% of adversarial samples on CIFAR10 and 94.46\% on SVHN with only 30.04 and 28.50 pruned models on average, respectively. Moreover, the FLOPs of a single model adopted in GGT is only 5.22\% of the model adopted in MMT.
\end{enumerate}

The rest of the paper is organized as follows. Next, we summarize the related works. In \autoref{Sec:RQ}, we discuss our research questions. In \autoref{Sec:GGT}, we introduce our method, followed by the results in \autoref{Sec:Exp}. Finally, we give the threats to validation in \autoref{Sec:Threats} and the paper is concluded in \autoref{Sec:Conclusion}.

\section{Related Works}
\label{Sec:Related work}
In this part, we give the related works on graph based DNN analysis, adversarial samples for DNN, and DNN testing for adversarial sample detection.

\subsection{Graph Structure and DNN}
In the real world, a lot of complex systems in biology, society, and technology can be described by graphs $G$ composed of a node set $V$ and an edge set $E$~\cite{chen2020graph}. Typical examples are brain network~\cite{fornito2016fundamentals}, communication network~\cite{akhtar2021graph}, and worldwide web~\cite{tan2020graph}, where each node represents a neuron, a person, or a web page, respectively, and the edges are the connections between each kind of these objects, which facilitate their information exchange.


DNN consists of layers of neurons and the connections between them, with its structure naturally captured by a graph, where each neuron is connected with those in the former and the next layers, with the extracted feature transmitted through edges. DNN architecture and its performance are highly correlated, which is widely recognized in the ML community~\cite{szegedy2015going,he2016deep}. Recently, You et al. ~\cite{you2020graph} established the relationship between DNN architecture and its performance using a relational graph. They argue that the directed data flow is too complex to model, and thus it is more reasonable to focus on the tractable message exchange.
\begin{figure} [!t]
\centering
\includegraphics[width=0.45\textwidth]{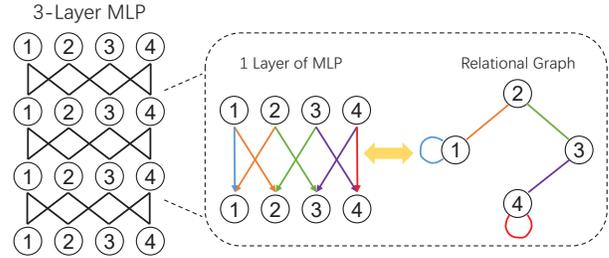}
\caption{Relational graph representation of a 3-layer MLP.}
\label{Fig:relational_graph}
\end{figure}

As shown in Fig.~\ref{Fig:relational_graph}, a 3-layer Multi-Layer Perceptron (MLP) with fixed layer width can be represented by a 4-node relational graph. Assume that each node $v$ has a node feature $x_v$, the edges between node 1 on the previous layer and nodes 1 and 2 on the next layer mean there are message transmissions from the previous layer to the next layer, which is defined by a message function $f(\cdot)$. The input of a message function is the previous node feature and the output is the updated node feature. At the inference stage, MLP computes the input feature layer by layer. The corresponding process in the relational graph is that message functions transform node features, then their outputs are aggregated at each node by an aggregation function $A(\cdot)$. For an $n$-layer MLP, the $r$-th round message exchange can be described as
\begin{equation}
x_v^{r+1} = A^{(r)}(f_v^{(r)}(x_u^{(r)}), \forall u\in N(v)),
\end{equation}
where $u$ and $v$ reppresent two nodes in the relational graph, $N(v)$ is the neighborhood of $v$, $x_u^{(r)}$ is the input node feature of $u$, and $x_v^{r+1}$ is the output node feature of $v$.

The relational graph can also be applied to more complex DNN architectures, including Convolution Neural Networks~\cite{krizhevsky2017imagenet} (CNNs) and residual connections~\cite{he2016deep}. 

\subsection{Adversarial Samples for DNN}
\label{Subsec:adversarial}
Adversarial attacks exist in a variety of deep learning application scenarios, e.g., image classification~\cite{moosavi2016deepfool}, link prediction in social networks~\cite{yu2019target}, and natural language processing~\cite{zhang2020adversarial}. This work focuses on adversarial samples in image classification. 
Adversarial attacks can be divided into black-box attacks and white-box attacks, with respect to their knowledge of the DNN model information. White-box attacks grab the whole information of the target model, including model structure and parameters.
Black-box attacks only know the input and output of the DNN model, and thus usually need greater perturbation to make a successful attack.
We will introduce several most commonly used adversarial attack methods, including 4 white-box attacks (FGSM, JSMA, CW, and Deepfool) and 2 black-box attacks (Local Search Attack and One Pixel Attack).

\subsubsection{FGSM}
Fast Gradient Sign Method (FGSM) is proposed by Goodfellow et al.~\cite{goodfellow2014explaining}. They found the derivative of the model to the input, then used a sign function to get the gradient direction. The perturbation is obtained by multiplying by one step in the sign direction and the final adversarial sample $\hat{x}$ is
\begin{equation}
\label{Eq:FGSM}
\hat{x} = x + \varepsilon sign(\nabla_xJ(x,y)),
\end{equation}
where $x$ is the benign sample, $\varepsilon$ is the step, $J$ is the loss function of the trained model, and $y$ is the ground truth label of $x$. FGSM is ``fast'' because it only updates the perturbation once, and the result is not guaranteed to be minimal.

\subsubsection{JSMA}
Jacobian-based Saliency Map Attack~\cite{papernot2016limitations} (JSMA) is a kind of targeted attack that utilizes the adversarial saliency between the input feature and the output only by modifying a small number of input pixels. There are 3 steps in JSMA, calculating the forward derivative, calculating the adversarial saliency map, and adding perturbation. The forward derivative is obtained by deriving the output of the model's last layer for a given input. Then the attacker calculates a saliency map based on the Jacobian matrix that reflects the impact of different input features on the output. The final perturbation is added on the top-2 dominating features (2 pixels) of the input.

\subsubsection{CW}
Carlini \& Wagner Attack~\cite{carlini2017towards} (CW) is a kind of optimization based attack method. The idea is treating the input as a variable, train the input (add perturbation) with fixed model parameters, maximize the distance between the ground truth label and the model output, and minimize the distance between the target label and model output.
%
\subsubsection{Deepfool}
Deepfool Attack~\cite{moosavi2016deepfool} aims to find the shortest distance from the normal sample to the classification hyperplane and cross the boundary to generate adversarial samples. For two-classification problem, the object of perturbation $\mathbf{r}$ is
\begin{equation}
\begin{aligned}
& argmin_{\mathbf{r}}\|\mathbf{r} \|_2 \\
& s.t. \ sign(f(x_0+\mathbf{r}))\neq sign(f(x_0)),
\end{aligned}
\end{equation}
where $f(\cdot)$ is the model output on the classification hyperplane and $x_0$ is the normal sample. This objective function can be solved iteratively to obtain the smallest perturbation. For the multi-classification problem and the detailed derivation process, we refer readers to ~\cite{moosavi2016deepfool}.
\subsubsection{Local Search Attack}
Narodytska et al.~\cite{narodytska2017simple} proposed a simple black box attack with no internal knowledge of the target network. Their attack uses local search to construct an approximation of the model gradient, then uses it to guide the generation of perturbation. The object is to minimize the output confidence of benign sample $I$ on label $c$. There are mainly two steps in the local search. First, obtain the confidence $f(\tilde{I})$ with the perturbation added on the last iteration, where $\tilde{I}$ is the perturbed image, and sorts the confidence scores in descending order; Second, the perturbation is added depending on whether the attack is successful. This process will end after the specified number of iterations or the attack successes.
\subsubsection{One Pixel Attack}
One Pixel Attack achieves misclassification by only modifying one pixel of the input image~\cite{su2019one}. The object is
\begin{equation}
\begin{aligned}
& maximize \ f_{adv}(x+e(x))  \\
& subject \ to \ \| e(x) \|_0 \leq 1,
\end{aligned}
\end{equation}
where $x$ is the normal sample, $e(x)$ is the perturbation, and $f_{adv}$ is the model confidence on the target adversarial label. To obtain the best pixel location and perturbation amplitude, Su et al.~\cite{su2019one} used differential evolution. The perturbation is represented by a 5-element vector, which is the x-y axis coordinates and the disturbance amplitude of each RGB channel. We refer readers to ~\cite{su2019one} for details.

\subsection{Testing for Adversarial Sample Detection}
Testing for DNN models aims to find bugs (adversarial samples) or evaluate the quality of a test set, based on which we can fix the model and avoid vulnerabilities being exploited. Quite recently, DNN testing technologies are continuously developed for adversarial sample detection. Guo et al.~\cite{guo2020audee} proposed a testing framework to detect logical bugs, crashes, and Not-a-Number (NaN) errors in deep learning models. However, their work only involves software-level bugs, but cannot detect bugs (adversarial samples) hidden in the model structure and weights. Ma et al.~\cite{ma2019nic} found that adversarial samples have different activation patterns from normal ones, and thus introduced invariant checking for adversarial sample detection. Yin et al.~\cite{yin2019adversarial} thought that adversarial and normal samples can be distinguished in the input subspace, i.e., adversarial samples are closer to the decision boundary, based on which they trained a number of detectors using asymmetrical adversarial training. Wang et al.~\cite{wang2020dissector} proposed that for a normal input, a deep learning model should process it with increasing confidence. They dissected the middle layers of DNN to generate a set of sub-models whose prediction profile for the input can be used to determine whether the corresponding input is within the processing capacity of the model, then a sample is considered as normal if it is within the handling capabilities of the sub-models, otherwise, it is adversarial. Zhang et al.~\cite{zhang2020towards} investigated the uncertainty patterns between normal samples and adversarial samples, raised an automated testing technique to generate testing samples with diverse uncertainty patterns. Their work mainly focuses on providing a high quality testing dataset and evaluating existing defense techniques.

However, detecting DNN bugs is not easy, Carlini and Wagner~\cite{carlini2017adversarial} evaluated ten detection methods proposed in the past years on different threat models and found that six of them are significantly less effective under generic attacks. Moreover, with a white-box attack that designed for penetrating a given defense mechanism, half of the methods fail to provide robustness, three increase robustness slightly, and two only works on simple datasets. All of the evaluated methods are based on a single model, and their criteria can be easily added to the penalty of adversarial sample generation to bypass the detection mechanism. We argue that the testing methods based on multiple models, such as MMT, can be more robust in adversarial sample detection. In fact, MMT indeed achieves SOTA performance in detecting various adversarial samples, when it integrates enough mutated models. However, each coin has two sides, and the huge resource cost may hinder its wide application. This is the reason why we propose GGT in this paper, which is based on multiple pruned models, and much simpler than MMT, making it more applicable in reality.

\section{Research Questions}
\label{Sec:RQ}
When we use multiple models to detect adversarial samples, it is crucial to ensure that each individual model has competitive accuracy with the original one. Therefore, in MMT, the mutation rate is set to a very small value to make sure that the mutated models are valid, i.e., their performance is not significantly worse than the original model. Such a setting may result in a huge redundancy, leading to unnecessary resource consumption. On the other hand, recently, a lot of pruning methods are proposed to simplify DNN models without losing their effectiveness~\cite{lin2020hrank,liu2020autocompress}. An interesting approach is to use a relational graph to capture the structure of DNN~\cite{you2020graph}, and then prune DNN models under the guidance of sparse graph design. Traditionally, graph sparsity is always represented by the small AD, then we give the first research question, \textbf{RQ1}: \emph{Can we significantly reduce the complexity of DNN models with low accuracy loss, guided by the AD of the relational graph, so that we can use smaller pruned models to design our GGT for efficient adversarial sample detection?}

Generally, in network science, ASPL, defined as the minimum number of links required to connect any pair of nodes, is an important structural property, which is strongly correlated with a series of network dynamics~\cite{costa2007characterization}, e.g., small ASPL can reduce the communication cost and increase the ability of synchronization~\cite{xuan2009optimal}. As DNN structure is described by a relational graph, the training and testing of DNN can be considered as certain dynamics on the relational graph. Therefore, it is expected that the ASPL of a relational graph could also be correlated with the performance of DNN model, which leads to our second question, \textbf{RQ2}: \emph{What is the relationship between the performance of the pruned DNN and the ASPL of the relational graph? Can we adopt the DNN models with longer (or shorter) ASPL to design better GGT so as to improve the adversarial sample detection?}

\label{Sec:GGT}
\begin{figure*}[!t]
\centering
\includegraphics[width=.85\textwidth]{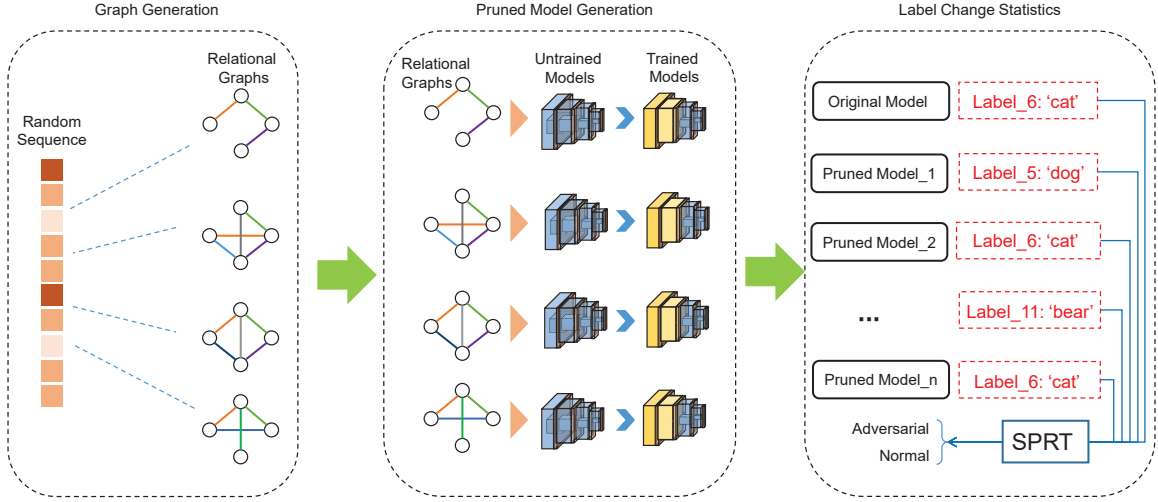}
\caption{The Design of Graph-Guided Testing.}
\label{Fig:method_diagram}
\end{figure*}

Based on information theory, a sparse graph can provide more structural information than a fully connected graph, i.e., a large number of graphs with different structure could be generated for a particular AD satisfying $k<N-1$, while only one fully connected graph can be generated when it is satisfied $k=N-1$, supposing the number of nodes $N$ in the graph is fixed. Such diversity can indeed benefit graph algorithms, such as node matching between networks~\cite{xuan2009node}. Therefore, it is expected that the integration of DNN models with sparse structure can benefit our GGT, i.e., we may need a smaller number of pruned models in our GGT to get better adversarial sample detection accuracy, compared with MMT. On the other hand, \emph{there is no such thing as a free lunch}, i.e., it may be argued that since a single pruned model in GGT has significantly smaller computational cost than the mutated model adopted in MMT, we may need more pruned models to detect adversarial samples with comparable accuracy. This leads to our third question, \textbf{RQ3}: \emph{Does the diversity of pruned DNN models, introduced by the sparsity of relational graphs, benefit our GGT for better detecting adversarial samples? Are smaller number of pruned models required in GGT to get comparable detection accuracy, compared with MMT?}


In MMT, various mutation operators are adopted to make the decision boundaries of the mutated models slightly different from the decision boundary of the original model. Therefore, MMT metrics favor identifying the adversarial samples that locate near the decision boundary, which have relatively low confidence in the model output~\cite{wen2020towards}. In other words, MMT may not be able to detect the high-confidence adversarial samples that are relatively far from the decision boundary. Our GGT, on the other hand, is based on the pruned models which may be significantly smaller than the original model, i.e., their decision boundaries could be more different. As a result, GGT may be more effective than MMT in detecting high-confidence adversarial samples. Thus it is interesting to give our last question, \textbf{RQ4}: \emph{Is the superiority of GGT over MMT even more significant on detecting high-confidence adversarial samples that are relatively far from the decision boundary?}

\section{Graph-Guided Testing}
\label{Sec:GGT}
Our Graph-Guided Testing (GGT) for adversarial sample detection consists of three steps, graph generation, pruned model generation, and label change statistics, as shown in Fig.~\ref{Fig:method_diagram}. Given a DNN model, we first generate diverse relational graphs with a certain number of nodes and edges according to the predefined AD. We then generate pruned DNN models based on the relational graphs, which are further integrated to detect adversarial samples.

The detection is based on the sensitivity of the pruned models to different kinds of samples. For a normal sample, pruned models tend to give consistent outputs; for an adversarial sample, there will be a variety of labels in the outputs. We use statistical hypothesis testing to capture such difference, and the threshold is determined by normal samples. Thus our method can be used to detect unknown adversarial samples.

\subsection{Graph Generation}
\label{Subsec:graph_gen}
There are a lot of characteristics, such as degree distribution, Average Degree (AD), Average Shortest Path Length (ASPL), average cluster coefficient, etc., to represent the graph structure. In this work, we mainly focus on undirected graphs, and simply set that each node has the same degree since the degree of a normal DNN internal neurons are equal.
In this case, AD can be used to control the sparsity of relational graphs, so as to determine the pruning rate of the corresponding DNN model. We explore relational graphs with a wide range of ASPL to better answer \textbf{RQ2}.

According to the definition of relational graph~\cite{you2020graph}, we define an undirected graph $G=(V, E)$ by a node set $V={v_1, ..., v_n}$ and an edge set $E\subseteq \{(v_i, v_j)|v_i, v_j\in V\}$. Each node $v_i$ has an associated feature $x_i$, which is a tensor representing multi-channel data. The number of nodes $N$ must be less equal than the number of channels in the DNN model. Suppose there are $M$ links in a relational graph with $N$ nodes (self-connection is not considered here), the AD of the graph is calculated by
\begin{equation}
k=\frac{2M}{N}.
\end{equation}
The \emph{pruning rate} (or \emph{sparsity ratio}) of the DNN model is then defined as
\begin{equation}
\theta=1-\frac{k}{N}.
\end{equation}


\begin{figure} [!ht]
\centering
\includegraphics[width=0.35\textwidth]{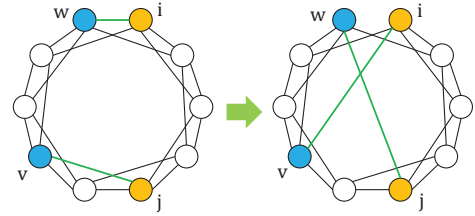}
\caption{Optimization process of regulating ASPL.}
\label{Fig:APL_opt}
\end{figure}

\begin{algorithm}[ht]\scriptsize
    \renewcommand{\algorithmicrequire}{\textbf{Input:}}
    \renewcommand{\algorithmicensure}{\textbf{Output:}}
    \caption{ASPL regulation of relational graph}
    \label{Alg:graph_gen}
    \begin{algorithmic}[1]
        \REQUIRE
        A randomly generated graph $G_i^{(0)}$ with fixed number of nodes $N$ and initial ASPL $D_i$, maximum number of exchanges $m$, target ASPL boundary $\left[L_l,L_h\right)$, and exchange counter $n$.
        \ENSURE A graph $G_o$ within required ASPL range $\left[L_l,L_h\right)$.
        \REPEAT
            \STATE Select two nodes $v_i$ and $v_j$ with their respective neighbors $v_w$ and $v_v$ that $v_i$, $v_j$, $v_w$, and $v_v$ must be 4 different nodes, $n=0$;
            \STATE Exchange $v_i$ and $v_j$ neighbors, $n=n+1$, obtain $G_i^{(n)}$;
            \IF {$G$ is not connected}
                \STATE Go to Step.2;
            \ENDIF
            \STATE Calculate the ASPL of $G_i^{(n)}$ and $G_i^{(n-1)}$ with $D=\frac{2}{N\times (N-1)}\sum_{i=2}^N\sum_{j=1}^{i-1}d(v_i,v_j)$;

            \IF {$D_i\leq L_l$ and $D(G_i^{(n)}) > D(G_i^{(n-1)})$}
                \STATE Save the exchange;
            \ELSIF{$D_i > L_h$ and $D(G_i^{(n)}) < D(G_i^{(n-1)})$}
                \STATE Save the exchange;
            \ELSE
                \STATE Reject the exchange;
            \ENDIF

            \IF {$L_l \leq D(G_i^{(n)}) < L_h$}
                \RETURN graph $G_i^{(n)}$;
            \ENDIF
        \UNTIL{$n > m$}
        \RETURN fail to generate required graph;
    \end{algorithmic}
\end{algorithm}

For the generated graph, we further regulate its ASPL to obtain a series of relational graphs with different ASPL. The object of regulating ASPL is given by
\begin{equation}
D=\frac{2}{N\times(N-1)}\sum_{i=2}^N\sum_{j=1}^{i-1}d(i,j),
\end{equation}
where $N$ is the total number of nodes in the graph, $d(\cdot)$ is the distance function that measures the shortest path length between nodes $v_i$ and $v_j$. The regulation of ASPL follows the rewiring process proposed by Xuan et al.~\cite{xuan2009optimal}. The details are listed in Algorithm~\ref{Alg:graph_gen}. As shown in Fig.~\ref{Fig:APL_opt}, we first randomly choose two nodes in the graph, and then exchange their neighbors. Note that the selected four nodes must be different. For example, in Fig.~\ref{Fig:APL_opt}, node $v_i$, its neighbor node $v_w$, node $v_j$, and its neighbor $v_v$ are selected. We delete the edges between $v_i$ and $v_w$, $v_j$ and $v_v$, and then create new edges to connect $v_i$ and $v_v$, and $v_j$ and $v_w$. This ensures the degree of each node fixed during the regulation. After new edges are created, we make the following check: If the graph is connected after rewiring and the current ASPL $D$ is closer to the target ASPL interval $\left[L_l,L_h\right)$, we accept the update, otherwise, we reject it, and then return to the node selection step. Once the current ASPL satisfies $D\in\left[L_l,L_h\right)$, we record the relational graph. Once the maximum number of rewiring times is achieved, we stop the rewiring process. Note that, we randomly initialize a large number of relational graphs (all the nodes have equal degrees) and repeat the above process for each of them, in order to get enough relational graphs with their ASPL falling into various predefined intervals.


\subsection{Pruned Model Generation}
\begin{figure} [!t]
\centering
\includegraphics[width=0.35\textwidth]{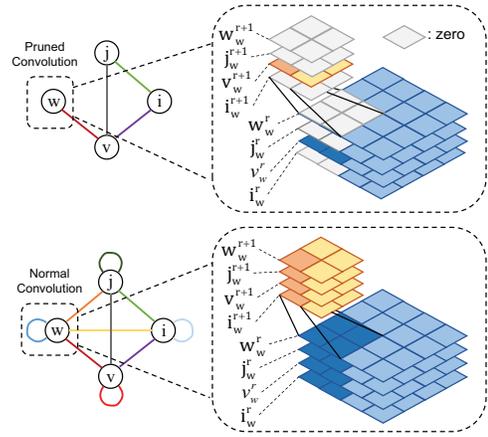}
\caption{Graph mapping in different situations.}
\label{Fig:graph_map}
\end{figure}

Pruned DNN models are created using the adjacency matrix of the above generated relational graph. For a graph $G=\{V,E\}$ with $N$ nodes, the elements of its adjacency matrix $A_{N\times{N}}$ indicate whether pairs of vertices are adjacent or not in the graph. If there is an edge between nodes $v_i$ and $v_j$, i.e., there is message exchange between them, we have $A[i][j]=1$, otherwise $A[i][j]=0$. Considering a DNN with complex components, such as convolution layers and pooling layers, we define the node feature as a tensor $\bm{X}_i$ and the message exchange as convolutional operator, which is defined as
\begin{equation}
\bm{X}_i^{(r+1)}=AGG(\sum_{j\in N(i)}(A[i][j]\times \bm{W}_{i,j}^{(r)}) * \bm{X}_{j}^{(r)}),
\end{equation}
where $AGG(\cdot)$ is an aggregation function, $*$ is the convolutional operator, $N(i)$ is the neighbors of node $v_i$, and $v_r$ represents the $r$th round of message exchange. A Pooling can be regarded as a special convolution, a pooling message exchange can be expressed as
\begin{equation}
\bm{X}_i^{(r+1)}=AGG(max_{m}(\bm{X}_{j}^{(r)})),
\end{equation}
where $max(\cdot)$ is the max pooling operator and $m$ is the pooling kernel size.
As shown in Fig.~\ref{Fig:graph_map}, in a relational graph, a round of message exchange means the extracted feature is passed to the next layer. In a normal convolutional layer, the convolution kernel will traverse the entire feature map, and in the corresponding relational map, it is expressed as one node is linked to all the other nodes. There could be redundant edges that exist in this kind of fully connected relational graph, and reducing the redundancy in the graph may in turn prune the DNN model. Using the generated sparse relational graph, we obtain a pruned model after mapping. The pruning is achieved by using a masking matrix multiplying the original convolution weight, where the reserved weight is multiplied by 1, and the removed weight is multiplied by 0. After pruning, the number of feature maps is constant, but the values of those feature maps corresponding to the masked weights are all equal to 0.

Different layers of a DNN model are generally different in size. We allow that the same node in different layers or within the same layer has different dimensions. Specifically, suppose one layer has $n_c$ channels, and the total number of nodes in the relational graph is $N$. Then there are ($n_c \ \text{mod} \ {N}$) nodes that have $\lfloor n_c/N\rfloor +1$ channels. According to the number of channels that a node is assigned, we create a mask matrix as described above, and multiply it by the weights corresponding to the channels. That is, the weights of the corresponding edges in the relational graph are retained.

We will retrain these pruned models to improve their accuracy. Note that the masked weights can be removed in post-process, thus the actual parameters and computational cost of the pruned models can be significantly reduced. Here, we argue that, when we use the retrained pruned models to detect adversarial samples, it's testing time, rather than training time, that really matters in the real application. Therefore, it is quite important to simplify DNN models by using our graph-guided pruning technology to reduce the memory and computational cost in the online detection phase. 

\subsection{Label Change Statistics}
\label{Subsec:lcr}
\begin{figure} [!t]
\centering
\includegraphics[width=0.33\textwidth]{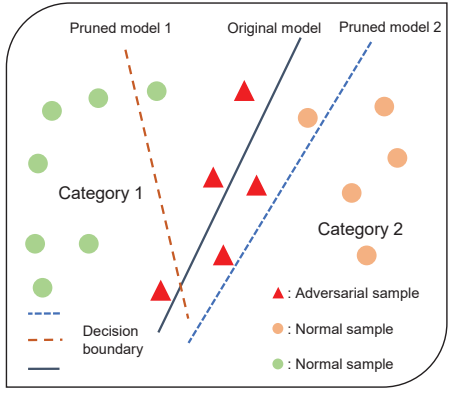}
\caption{Decision boundary of the original DNN model and the pruned models generated from relational graphs.}
\label{Fig:boundary}
\end{figure}

As shown in Fig.~\ref{Fig:boundary}, adversarial samples are generally near the original model decision boundary, while pruned models have different boundaries from the original model. On the one hand, pruned models have a small probability to give wrong output for some normal samples (because of pruning and retraining); on the other hand, their boundaries may also be more sensitive to adversarial samples. Because the model decision boundaries are complex, and an adversarial sample that crosses the boundary of the original model may also cross different boundaries of the pruned models, i.e., outputting multiple different labels. We thus can use Label Change Rate (LCR) to measure the diversity~\cite{wang2019adversarial} of the pruned models' outputs, which is defined as
\begin{equation}
\label{Eq:lcr}
\eta = \frac{\sum_{s\in S}E(f(x),s(x))}{|S|},
\end{equation}
where $S$ is the set of pruned models, $x$ is the input, $f(x)$ is the output of the original model, $s(x)$ is the output of the pruned model, $|S|$ is the size of set $S$,
and $E(\cdot)$ is defined as
\begin{equation}
E(x,y)=
\begin{cases}
0 & \ \text{if} \ x=y, \\
1 & \ \text{otherwise},
\end{cases}
\end{equation}
which counts the number of times that the classification results in the pruned models are different from that of the original model. We assume that, given a certain number of pruned models, normal samples and adversarial samples will have significantly different scores. According to the experimental results, we then set a threshold to distinguish adversarial and normal samples.

Specifically, we use the Sequential Probability Ratio Testing~\cite{wald2004sequential} (SPRT) to determine whether an input is adversarial dynamically. It is usually faster than using a fixed number of pruned models. The SPRT probability ratio is calculated by
\begin{equation}
pr = \frac{p_1^z(1-p_1)^{n-z}}{p_0^z(1-p_0)^{n-z}},\label{LCR}
\end{equation}
where $n$ is the total number of used pruned models, $z$ is the number of pruned models that output different labels, $p_0=\eta_h+\sigma$, and $p_1=\eta_h-\sigma$. $\eta_h$ is a threshold, which is determined by the LCR of normal samples. We calculate the Area Under the Receiver Operating Characteristic (AUROC) to determine whether normal and adversarial samples can be distinguished by $\eta_h$. That is, we get the LCR values of a set of normal samples and a set of adversarial samples using Eq.~(\ref{Eq:lcr}), and then calculate AUROC for every possible $\eta_h$. In the AUROC, the x-axis and y-axis represent the true positive rate and false positive rate obtained by using the threshold $\eta_h$, respectively. The closer AUROC is to 1, the better the threshold is. $\sigma$ is a relax scale, which means neither hypothesis (normal or adversarial) can be denied in the region $(\eta_h-\sigma, \eta_h+\sigma)$. Then, we calculate the \emph{deny} LCR $DL$, and \emph{accept} LCR $AL$ using
\begin{equation}
\begin{split}
  DL&=log_{e}\frac{\beta}{1-\alpha}, \\
  AL&=log_{e}\frac{1-\beta}{\alpha},
\end{split}
\end{equation}
where $\alpha$ and $\beta$ denote the probability of false positive (a normal sample is misclassified as an adversarial sample) and false negative (an adversarial sample is misclassified as a normal sample), respectively. During the testing, the input will be sent to the original model as well as the pruned models. Then, we can get the dynamic LCR based on Eq.~(\ref{LCR}), denoted by $pr$, which is compared to the \emph{deny} LCR and \emph{accept} LCR. If $pr\leq DL$, the input is considered as an adversarial sample, while if $pr\geq AL$, it is considered as a normal one. The whole process is shown in Algorithm~\ref{Alg:SPRT}.

\begin{algorithm}[ht]\scriptsize
    \renewcommand{\algorithmicrequire}{\textbf{Input:}}
    \renewcommand{\algorithmicensure}{\textbf{Output:}}
    \caption{Adversarial sample detection using SPRT}
    \label{Alg:SPRT}
    \begin{algorithmic}[1]
        \REQUIRE
        Relational graph $G$,
        input $x$,
        original model $f$,
        threshold $\eta_h$,
        relax region $\sigma$,
        deny LCR $DL$,
        accept LCR $AL$,
        maximum number of models $m$,
        counter $n$ for used pruned models, and
        counter $z$ for different outputs of pruned models.
        \ENSURE Whether a sample is adversarial.
        \WHILE{$n<m$}
            \STATE Generate and retrain a pruned model $s$ with a relational graph $G$;
            \STATE $n = n + 1$;
            \IF {$s(x) \neq f(x)$}
                \STATE $z = z + 1$;
                \STATE Calculate $pr = \frac{p_1^z(1-p_1)^{n-z}}{p_0^z(1-p_0)^{n-z}}$, where $p_0=\eta_h+\sigma$, and $p_1=\eta_h-\sigma$;
                \IF{$pr\leq DL$}
                    \RETURN $x$ is a adversarial sample;
                \ENDIF
                \IF{$pr\geq AL$}
                    \RETURN $x$ is a normal sample;
                \ENDIF
            \ENDIF
        \ENDWHILE
    \end{algorithmic}
\end{algorithm}

\section{Experiments}
\label{Sec:Exp}
We implemented our GGT for adversarial sample detection with Pytorch (\textit{version} 1.6.0) based on Python (\textit{version} 3.7).
\subsection{Experiment Settings}
\subsubsection{Datasets and Models}
We evaluate our approach on two image datasets: CIFAR10 and SVHN. They are widely used in the evaluation of DNN testing frameworks~\cite{wang2019adversarial,xie2019deephunter,ma2018deepmutation}. The former consists of 50,000 images for training and 10,000 images for testing, while these numbers for the latter are 73,257 and 26,032, respectively. Note that here we do not use the MNIST dataset, since the adopted model for this dataset (e.g., LeNet) is too small, and thus the diversity of the relational graph could be largely limited. The image size of both datasets is $32\times32\times3$. ResNet18 and VGG16 are adopted for CIFAR10 and SVHN, with their accuracy equal to 93.03\% and 95.63\%, respectively. These models are independently trained in our experiment without fine adjustment of hyperparameters, thus the accuracy will fluctuate within a small range compared with the public models. GGT and MMT both use SPRT to detect adversarial samples, the number of used mutated or pruned models varies for different samples. For the parameters in SPRT, we set both $\alpha$ and $\beta$ equal to 0.05, and the relax scale $\sigma$ to 10\% of the threshold $\eta_h$. And we set the maximum number of models to 100 for both methods.

\subsubsection{Graph-Guided Pruned Models Generation}
For CIFAR10 and SVHN datasets, we set the number of nodes in each relational graph to 64 and use ASPL to guide our relational graph generation. We construct the adjacency matrix using the \textit{random\_degree\_sequence\_graph} function from the Python networkx package (\textit{version} 2.4), The parameter \textit{tries} is set to 1,000,000 and other parameters by default, which is empirically sufficient for successfully generating a valid adjacency matrix of a connected graph, where there must be at least one path connecting each pair of nodes. The AD, which determines the pruning rate of the generated DNN model, will affect the range of ASPL (too high or too low AD can not make ASPL reach the required length). We thus set the AD $k=3$ with the corresponding \emph{pruning rate} (or \emph{sparsity ratio}) equal to 95.3\% (1-3/64), under which the accuracy drop does not exceed 5\%. The ASPL ranges from 3 to 15 and is divided with a step of 2 (3-5, 5-7, 7-9, 9-11, 11-13, 13-15) to explore the relationship between adversarial sample detection accuracy and ASPL. There are 100 graphs for each segment. All the pruned models are retrained with accuracy at least equal to 85.34\% and 91.35\%, for CIFAR10 and SVHN, respectively.



\subsubsection{Adversarial Sample Generation}
We test GGT on detecting adversarial samples generated by six typical adversarial attack methods described in Sec.~\ref{Subsec:adversarial}, including 4 white-box and 2 black-box, and meanwhile those wrongly labeled samples are also considered as a kind of adversarial samples. The parameters for each attack are summarized as follows:
\begin{enumerate}
\item FGSM: the scale of perturbation is 0.03;
\item JSMA: the maximum distortion is 12\%;
\item CW: adopt L2 attack, the scale coefficient is 0.6 and the iteration number is 1,000;
\item Deepfool (DF): the maximum number of iterations is 50 and the termination criterion is 0.02;
\item One Pixel Attack (OP): the number of pixels for modification is 3 (in order to ensure that enough successful samples are generated) and the differential algorithm runs with a population size of 400 and a max iteration count of 100;
\item Local Search Attack (LS): the pixel complexity is 1, the perturbation value is 1.5, the half side length of the neighborhood square is 5, the number of pixels perturbed at each round is 5 and the threshold for k-misclassification is 1;
\item Wrongly Labeled (WL): the original model accuracy and test set size determine the number of WL samples.
\end{enumerate}

We choose the adversarial samples with confidence higher than 0.9 to perform the high-confidence attacks, regardless of the attack method. In addition to adversarial samples generated by these methods, as usual, we also regard the samples in the test set which are wrongly labeled by the original model as adversarial samples. We randomly select a certain number of samples from each kind of adversarial samples for evaluation, as presented in Table~\ref{Tab:adv_samples}.
\begin{table}[t]\scriptsize
	\caption{Number of generated samples.}
	\label{Tab:adv_samples}
    \centering
	\begin{tabular}{|c|c|c|c|c|c|c|c|c|}
		\hline
		 & Normal & WL & FGSM & JSMA & CW & DF & OP & LS  \tabularnewline
         \hline
         CIFAR10 & 1000 & 697 & 1000 & 1000 & 572 & 1000 & 714 & 647  \tabularnewline
         \hline
         SVHN & 1000 & 1000 & 1000 & 1000 & 1000 & 1000 & 196 & 309  \tabularnewline
         \hline
	\end{tabular}
\end{table}

\subsection{Evaluation Metrics}
We evaluate GGT in the following three ways:
\begin{enumerate}
\item Diversity Score Difference (DSD): the key to our GGT method is that normal sample and adversarial sample have significantly different LCR which is calculated by Eq.~(\ref{Eq:lcr}). Here, we define DSD as
\begin{equation}
d_l=\frac{l_a}{l_n}, \label{DSD}
\end{equation}
where $l_a$ and $l_n$ are the average LCR of adversarial and normal samples, respectively.
\item AUROC: GGT works based on the diversity score of normal samples. In order to select the best threshold and see how well it works, we calculate the Area Under the ROC (AUROC) curve to demonstrate whether the diversity score is an appropriate feature or not (the closer AUROC is to 1, the better the threshold is).
\item Accuracy of detection and the number of pruned models used: 
The higher the accuracy, the better our GGT is; the fewer pruned models needed for detection, the fewer resources it occupies.
\end{enumerate}
In order to better demonstrate the performance of our method, we compare it with the MMT method proposed by Wang et al.~\cite{wang2019adversarial} as the SOTA adversarial detection algorithm with a similar mechanism. They use four mutation operators to generate mutated models, i.e., Gaussian Fuzzing (GF), Weight Shuffling (WS), Neuron Switch (NS), and Neuron Activation Inverse (NAI), then distinguish adversarial samples by the statistical label change rate of mutated models. This method randomly mutates a trained model and the mutated models are not guaranteed to be accurate. Therefore, appropriate models should be selected from a bunch of candidates. 

\subsection{Results}
\textbf{Answer1: }
\emph{Based on our graph-guided mechanism, the DNN model complexity can be significantly reduced without much accuracy loss. Therefore, the much smaller pruned models can be used for subsequent  adversarial detection.}

\begin{figure} [!t]
\centering
\includegraphics[width=0.4\textwidth]{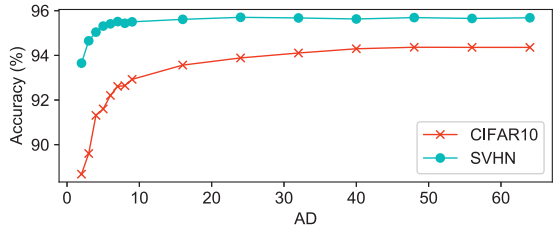}
\caption{Relationship between AD of the relational graph and the classification accuracy of the pruned DNN model.}
\label{Fig:adv_acc}
\end{figure}

To answer the RQ1, we first generate the graphs of different AD varied in $K$=\{2,3,4,5,6,7,8,9,16,24,32,40,48,56,64\}, with the degree of each node in a graph is set the same, equal to $k\in{K}$. For each $k$, we randomly create 5 graphs, generate their corresponding DNNs, train these DNNs on CIFAR10 and SVHN, and record the average testing accuracy. 
The relationship between the average accuracy of DNNs and the AD of the relational graphs is shown in Fig.~\ref{Fig:adv_acc}, where we notice that the accuracy of DNN indeed increases as the relational graph gets denser. Such trend is more significant when AD increases from 2 to 8, while the accuracy does not change much as AD further increases from 8 to 64. Moreover, the decrease of the accuracy is not much even when the relational graph gets very sparse, e.g., $k=2$ (with the pruning rate equal to 1-2/64$\approx$97\%). In this case, about 97\% of parameters can be pruned out from the original DNN, with the accuracy loss only equals to 4.35\% and 2.00\% on CIFAR10 and SVHN, respectively. As suggested in~\cite{wang2019adversarial}, the mutated models of accuracy over 90\% of the original model could be chosen for adversarial detection. Therefore we argue that all the graph-guided pruned models, with the relational graph keeping connected (no matter what AD is), can be adopted for adversarial detection, due to their relatively low accuracy loss. To improve the detection efficiency, and study the effect of ASPL on the classification accuracy of the pruned model, we set $k=3$ in the rest of the paper, since in this case, it is easier to get relational graphs of a wide range of ASPL.

\begin{figure} [!t]
\centering
\includegraphics[width=0.4\textwidth]{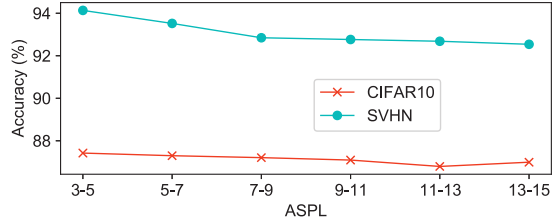}
\caption{Relationship between ASPL of the relational graph and the classification accuracy of the pruned DNN model.}
\label{Fig:aspl_acc}
\end{figure}

\textbf{Answer2: }
\emph{The shorter the ASPL of the relational graph is, the higher the classification accuracy of the pruned model. Meanwhile, the pruned models of shorter ASPL are better candidates to design GGT, since they can better capture the difference between adversarial samples and normal samples.}

We record the mean classification accuracy for each set of models with AD $k=3$, as shown in Fig.~\ref{Fig:aspl_acc}. As we can see, the accuracy slightly decreases as ASPL increases for both datasets. One possible reason is that the graph with a larger ASPL has a longer information transmission path between neurons, making it more difficult to extract effective features. 

\begin{figure}[!t]
\centering
\includegraphics[width=0.4\textwidth]{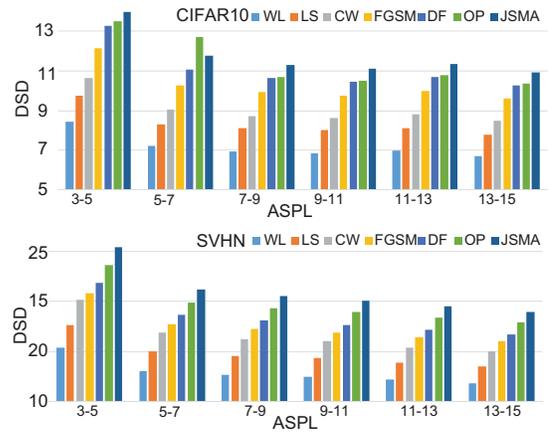}
\caption{DSD of GGT (with various ASPL).}
\label{Fig:diversity}
\end{figure}

To further explore the effect of ASPL on adversarial detection, we calculate the DSD of adversarial samples using 6 groups of pruned models with different ASPL, i.e., the average LCR of adversarial samples divided by that of normal samples, as represented by Eq.~(\ref{DSD}). Note that, as usual, we treat normal samples that wrongly labeled (WL) by the original model as a kind of adversarial sample~\cite{wang2019adversarial}, and the attacker will not add any perturbation. The results are summarized in Fig.~\ref{Fig:diversity}. In summary, for all the 6 groups of pruned models, adversarial samples have much higher LCR than normal samples (more than 5 times for CIFAR10 and 10 times for SVHN), indicating that these pruned models are highly sensitive to adversarial samples. Compared with other ASPLs, the pruned models with ASPL 3-5 show the largest DSD between normal samples and adversarial samples and thus are better candidates to design GGT for adversarial sample detection.

\textbf{Answer3: }
\emph{GGT outperforms MMT in both efficiency and effectiveness, i.e., GGT achieves significantly higher detection accuracy with much less DNN models, while the single pruned model in GGT is even much smaller than the single mutated model in MMT.}

\begin{figure} [!t]
\centering
\includegraphics[width=0.4\textwidth]{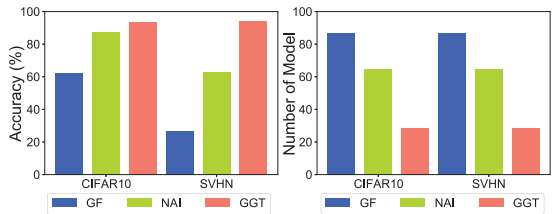}
\caption{Detection accuracy and the number of used models in GGT and MMT.}
\label{Fig:comparison}
\end{figure}

Here, we compare GGT with MMT under the same conditions. The number of the maximum used models is limited to 100 and other parameters of MMT are set to default according to the released code. Note that in the original paper of MMT, the model number limitation is set to 500, which we think is too huge for real applications~\cite{wang2019adversarial}. The mutation rate we use for MMT is 0.007 for both datasets to obtain the best performance. Our baseline is GF and NAI mutation operators, which are the best performers among their proposed methods. The adversarial sample detection accuracy and the number of used models are shown in Fig.~\ref{Fig:comparison}. It can be seen that GGT uses only 30.04 and 28.50 models on average considering all the attacks, while MMT needs 51.12 and 77.16 models, on CIFAR10 and SVHN, respectively. Note that the single pruned model in GGT is much smaller than the single mutated model in MMT. Such result suggests that GGT is much more efficient than MMT. Meanwhile, GGT achieves an average detection accuracy of 93.61\% and 94.46\%, on CIFAR10 and SVHN, while MMT only has 74.74\% and 44.79\%, respectively, indicating that GGT is also more effective than MMT in detecting adversarial samples.

%

In particular, we also calculate the AUROC score using LCR as the key feature to distinguish whether the input is an adversarial sample or not. The AUROC results are summarized in Table~\ref{Tab:AUROC_all}, with the best results marked in bold. We find that GGT has a generally higher AUROC than MMT, while the pruned models with ASPL 3-5 behave the best, in most cases, meaning that this set of pruned models have a good trade-off between true positives and false positives in detecting adversarial samples. This finding is in line with our expectation since this group of pruned models are of higher classification accuracy and can better capture the difference between adversarial samples and normal samples, as indicated by the results of RQ2.

\begin{table}[t]\scriptsize
	\caption{AUROC results (\%) for GGT and MMT.}
	\label{Tab:AUROC_all}
    \centering
	\begin{tabular} {|c|c|c|c|c|c|c|c|c|}
		\hline
          \multirow{2}{*}{Attack} &  \multicolumn{2}{c|}{MMT} & \multicolumn{6}{c|}{GGT with different ASPL} \tabularnewline
         \cline{2-9}
           ~  &   GF &  NAI &  3-5  &  5-7  &  7-9 &  9-11 & 11-13  & 13-15      \tabularnewline
         \hline
         \multicolumn{9}{|c|}{CIFAR10}  \tabularnewline
          \hline
          FGSM & 83.89 & 92.14 & \textbf{98.35} & 98.10 & 98.09 & 98.15 & 98.20 & 98.05 \tabularnewline
         \cline{1-9}
          JSMA            & 97.23 & 97.88 & \textbf{99.26} & 99.21 & 99.18 & 99.11 & 99.18 & 99.14 \tabularnewline
         \cline{1-9}
          CW                          & 85.85 & 91.70 & \textbf{97.29} & 96.97 & 96.82 & 96.91 & 97.04 & 96.80 \tabularnewline
         \cline{1-9}
          DF                          & 98.34 & 97.28 & \textbf{99.17} & 98.82 & 98.83 & 98.83 & 98.85 & 98.78  \tabularnewline
         \cline{1-9}
          OP                          & \textbf{99.95} & 99.77 & 99.35 & 99.03 & 98.98 & 99.00 & 99.02 & 98.99 \tabularnewline
         \cline{1-9}
          LS                          & 86.76 & 93.99 & \textbf{96.44} & 96.18 & 96.13 & 96.21 & 96.19 & 95.85 \tabularnewline
          \cline{1-9}
          WL                          & 85.92 & 92.38 & \textbf{93.90} & 93.35 & 93.05 & 93.06 & 93.32 & 93.01  \tabularnewline
         \hline
         \multicolumn{9}{|c|}{SVHN}  \tabularnewline
          \hline
          FGSM & 68.18 & 87.61    & 97.90 & 97.88 & 97.97 & \textbf{98.14} & 98.11 & 98.07 \tabularnewline
         \cline{1-9}
          JSMA                        & 83.76 & 96.25 & 99.31 & 99.27 & 99.27 & 99.28 & \textbf{99.32} & 99.24  \tabularnewline
         \cline{1-9}
          CW                          & 66.82 & 89.22 & \textbf{99.46} & 99.40 & 99.41 & 99.44 & 99.44 & 99.40  \tabularnewline
        \cline{1-9}
          DF                          & 83.47 & 96.59 & \textbf{99.74} & 99.68 & 99.66 & 99.69 & 99.68 & 99.66 \tabularnewline
         \cline{1-9}
          OP                          & 82.33 & 94.97 & \textbf{98.90} & 98.67 & 98.57 & 98.67 & 98.67 & 98.48  \tabularnewline
         \cline{1-9}
          LS                          & 79.27 & 93.30 & \textbf{98.42} & 98.27 & 98.26 & 98.28 & 98.25 & 98.11 \tabularnewline
        \cline{1-9}
          WL                          & 72.64 & 89.02 & 90.81 & 90.75 & 90.51 & 90.77 & \textbf{91.07} & 90.73   \tabularnewline
         \hline
	\end{tabular}
\end{table}

\textbf{Answer4: }
\emph{Considering only high-confidence adversarial samples, the adversarial detection accuracy increases 0.69 \% and drops 2.76\% on CIFAR10 and SVHN, respectively, for GGT; while it drops 16.04\% and 12.24\% on CIFAR10 and SVHN, respectively, for MMT.}

\begin{figure} [!ht]
\centering
\includegraphics[width=0.4\textwidth]{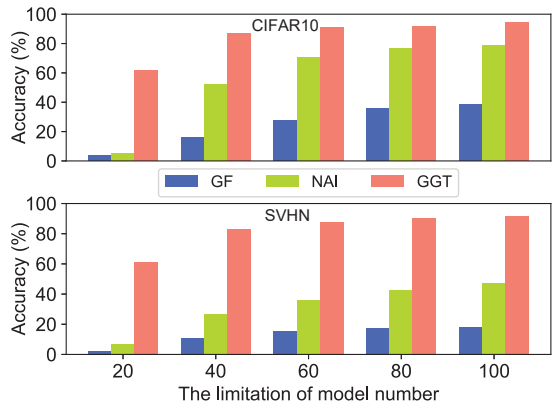}
\caption{Detection accuracy of high-confidence adversarial samples for various limitations of model number.}
\label{Fig:high}
\end{figure}


In this experiment, we choose 500 adversarial samples with confidence higher than 0.9 as high-confidence samples, regardless of which attack method is adopted. We find that, initially, MMT can detect 74.74\% and 44.79\% adversarial samples on CIFAR10 and SVHN, respectively, while these values decrease to 58.7\% and 32.55\% when detecting high-confidence adversarial samples, i.e., the overall detection accuracy drops almost 14.14\%. On the other hand, GGT can still detect 94.30\% and 91.70\% high-confidence adversarial samples (93.61\% and 94.46\% initially), i.e., the overall detection accuracy only drops 1.04\%. We further decrease the maximum number of models allowed in GGT and MMT, as shown in Fig.~\ref{Fig:high}, and found that the detection accuracy of MMT decreases quickly when only at most 20 models are allowed, i.e., it can barely detect any high-confidence adversarial samples, while GGT still detects 61.50\% and 61.40\% of them. Such results indicate that our GGT can detect both high-confidence adversarial samples and low-confidence adversarial samples with similar high accuracy, while MMT may lose its effectiveness in detecting high-confidence adversarial samples.

\section{Threats to Validity}
\label{Sec:Threats}
First, the graph-DNN mapping only works on CNN currently. Therefore, GGT may not be suitable for other kinds of DNN models. Nevertheless, CNN is still the mainstream deep learning method in computer vision, signal processing, and natural language processing, thus GGT is still valuable in these applications. We also focus on establishing graph-DNN mapping for other DNN models to make our GGT suitable for them in the near future.

Second, just like MMT, we only validate GGT on image data, while DNN models are also widely used in processing other types of data, e.g., time series. Considering that both GGT and MMT are based on the diversity of decision boundaries created by different DNN models, it is reasonable to believe that they can also be applied to other kinds of data.

Third, we only explore two graph characteristics, AD and ASPL, to generate pruned DNN models since AD determines the sparsity of a DNN model while ASPL is related to information exchanging efficiency. Indeed, our proposed GGT based on these two characteristics has both higher efficiency and effectiveness than MMT. There are certainly many other characteristics that could be further explored in the future.


\section{Conclusion}
\label{Sec:Conclusion}
In this paper, we establish the mapping between DNN architecture and relational graph and then prune a DNN model guided by the designed relational graph. Using this method, we can prune out more than 95\% parameters with only a small loss of accuracy, and find that the accuracy of the pruned model is negatively related to the ASPL of the relational graph. Based on these, we design Graph-Guided Testing (GGT) for adversarial sample detection, based on the pruned models with small AD and short ASPL in their corresponding relational graphs. The experimental results show that our GGT detects adversarial samples with 93.61\% and 94.46\% average accuracy, requiring 30.04 and 28.50 pruned models, on CIFAR10 and SVHN, respectively, much better than the state-of-the-art Model Mutation Testing (MMT) in both accuracy and resource consumption. In our future work, we will explore the structural space of DNN and design a better pruned DNN to further improve GGT. we will also extend GGT on more kinds of deep learning models, such as Recurrent Neural Network (RNN) and Graph Neural Network (GNN), and meanwhile adopt it to detect adversarial samples in various areas, such as signal processing and graph data mining.


\bibliographystyle{IEEEtran}
\bibliography{IEEEabrv,sample-base}

\end{document}